\documentclass{article}

\usepackage[final]{neurips_2022}
\usepackage{wrapfig}

\usepackage{multirow}

\usepackage{graphicx}
\usepackage[utf8]{inputenc} 
\usepackage[T1]{fontenc}    
\usepackage{hyperref}       
\usepackage{url}            
\usepackage{booktabs}       
\usepackage{amsfonts}       
\usepackage{nicefrac}       
\usepackage{microtype}      
\usepackage{xcolor}         

\usepackage{color}

\usepackage{algorithmicx}
\usepackage[ruled]{algorithm}
\usepackage{algpseudocode}
\usepackage{cancel}
\usepackage{makecell}
\usepackage[accsupp]{axessibility}  

\def\L{\mathcal{L}}

\def\our{HyperMAML}

\title{\our{}: Few-Shot Adaptation of Deep Models with Hypernetworks}

\author{%
    Marcin Przewięźlikowski$^{\dag,1,2,3}$ \thanks{
    Corresponding author: \texttt{marcin.przewiezlikowski@doctoral.uj.edu.pl}\\
    $\dag$ -- M.P. and P.P. contributed equally
    }
    \quad Przemysław Przybysz$^{\dag,1}$\\
    \textbf{Jacek Tabor}$^{1}$
    \quad \textbf{Maciej Zięba}$^{4,5}$
    \quad \textbf{Przemysław Spurek}$^{1,3}$\\
    $^1$ Jagiellonian University, Faculty of Mathematics and Computer Science\\
    $^2$ Jagiellonian University, Doctoral School of Exact and Natural Sciences\\
    $^3$ IDEAS NCBR\quad$^4$  Tooploox Ltd.\quad$^5$ Wroclaw University of Science and Technology
}

\begin{document}

\maketitle

\begin{abstract}

The aim of Few-Shot learning methods is to train models which can easily adapt to previously unseen tasks, based on small amounts of data. 
One of the most popular and elegant Few-Shot learning approaches is Model-Agnostic Meta-Learning (MAML). The main idea behind this method is to learn the general weights of the meta-model, which are further adapted to specific problems in a small number of gradient steps. However, the model's main limitation lies in the fact that the update procedure is realized by gradient-based optimisation. In consequence, MAML cannot always modify weights to the essential level in one or even a few gradient iterations. On the other hand, using many gradient steps results in a complex and time-consuming optimization procedure, which is hard to train in practice, and may lead to overfitting. In this paper, we propose \our{}, a novel generalization of MAML, where the training of the update procedure is also part of the model. Namely, in \our{}, instead of updating the weights with gradient descent, we use for this purpose a trainable Hypernetwork. Consequently, in this framework, the model can generate significant updates whose range is not limited to a fixed number of gradient steps. Experiments show that \our{} consistently outperforms MAML and performs comparably to other state-of-the-art techniques in a number of standard Few-Shot learning benchmarks. 
\end{abstract}

\section{Introduction}

In the typical Few-Shot learning setting, the aim is to adapt to new tasks under the assumption that only a few examples are given. As we know, people typically learn new tasks easily by using only a few training examples. On the contrary, a standard deep neural network must be trained on an extensive amount of data to obtain a similar accuracy.
Thus, the aim of Few-Shot learning models is to bring neural networks closer to the human brain's capabilities. The most famous and, in our opinion, the most elegant approach to Few-Shot learning is Model-Agnostic Meta-Learning (MAML) \citep{finn2017model}, where the model is trained to adapt universal weights to new Few-Shot learning tasks quickly.
It seems that the brain's neural networks can adapt to new tasks too, by applying the fact that during the process of evolution, some of its parts have developed universal weights 
which are easily adaptable to typical tasks we encounter in real life. Thus the idea behind MAML gives us a possible insight into the working of the brain. 

The fascinating factor of human intelligence is that the human learning process, although still not understood, is clearly not based on the gradient descent algorithm, as we cannot in general backpropagate the information \citep{lillicrap2020backpropagation,song2020can,whittington2019theories}. Thus, from the biological point of view, the main limitation of MAML is the fact that it uses the gradient descent method for weight updates. 
The main research problem that we set ourselves is whether one can modify MAML to be more biologically feasible, i.e. keep its ability to find universal weight but remove the necessity of using gradient-based update methods.

\begin{figure}[t]
	\centering
    \includegraphics[width=\textwidth]{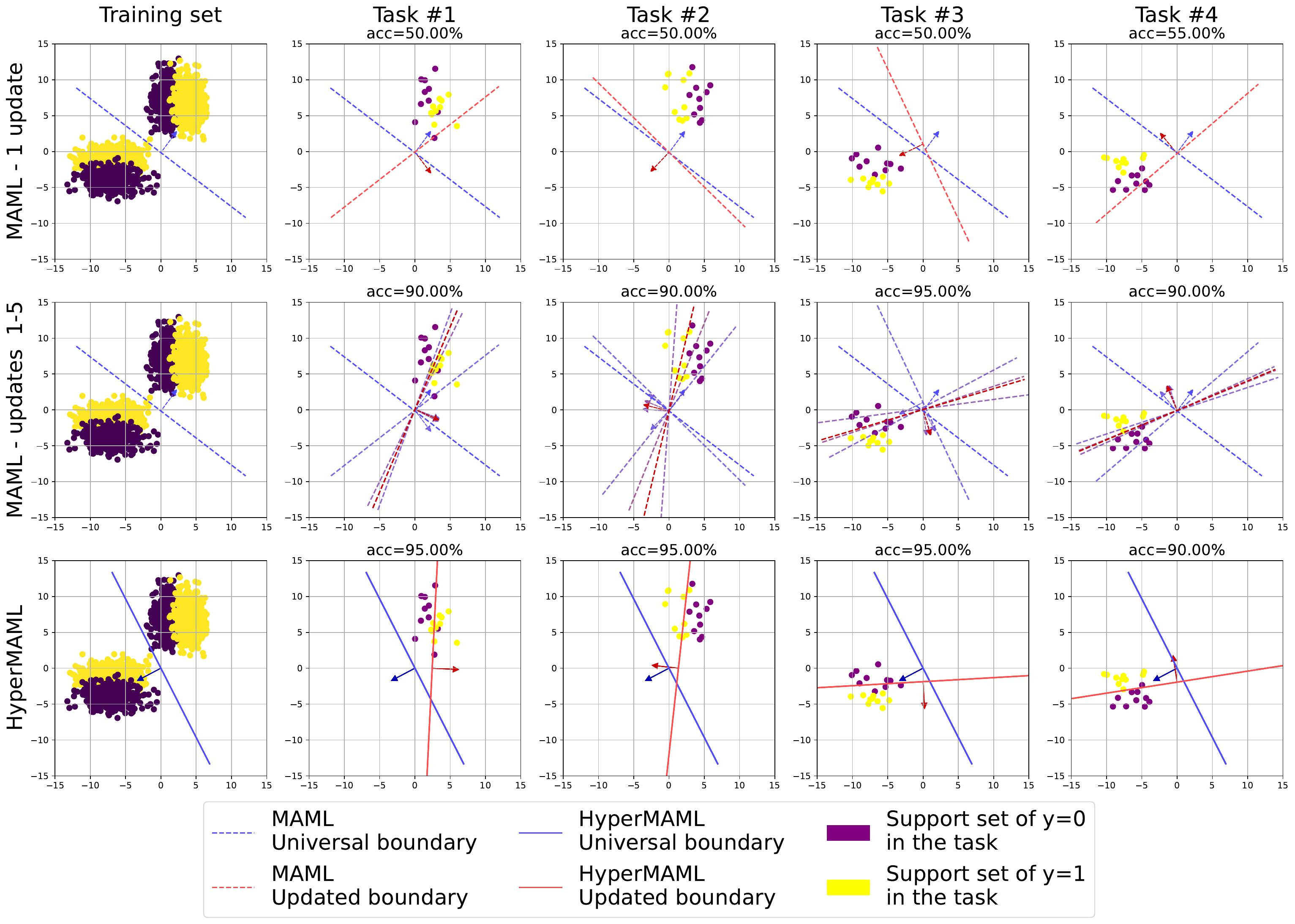}
    \caption{We consider a two-dimensional dataset consisting of four Gaussian data (the first column). In the Meta-Learning scenario, we produce a task that consists of samples from two horizontal or vertical ellipses with permuted labels (second to the fifth column).
    The MAML model cannot update its parameters for all four tasks with a single gradient step in the inner loop (see the first row). To reach a reasonable solution, it requires up to five gradient updates (see the second row).
    Our \our{} can solve the task by using the Hypernetwork paradigm. {Our method learns to perform only a single update, which nevertheless yields parameters optimal for given tasks.}}
    \label{fig:2D}
\end{figure}

We solve this problem by constructing  \our{}\footnote{We make our implementation available at \url{https://github.com/gmum/few-shot-hypernets-public}.}, a model which replaces the gradient optimization in the update of weights by trainable update procedure with the use of the Hypernetwork paradigm.
Hypernetworks, introduced in \citep{ha2016hypernetworks} are defined as neural models that generate weights for a separate target network solving a specific task. In our model, \our{}, the Hypernetwork aggregates the information from the support set and produces an update to the main model. Thanks to such an approach, we can create various types of updates that are not limited to a few gradient steps.

In practice, MAML works when there exist universal weights that are close enough to the optimal solution for each task. We visualize such a situation in the form of a simple 2D example, where a single gradient update fails to sufficiently adapt the model to a given task --  see Fig.~\ref{fig:2D}. We cannot effectively switch weight in one gradient step. On the other hand, when we use many gradient steps, we obtain a complex optimization procedure that uses an inner and outer loop. Such a procedure can be seen as second-order optimization, which is complex to train \citep{finn2017model}.
Contrary to MAML, \our{} does not need an inner loop in the optimization procedure, and consequently, we do not have second-order optimization. 
As a result, our algorithm obtains better results than the classical MAML algorithm and produces results comparable to other state-of-the-art algorithms.

The contributions of our work can be summarized as follows:
\begin{itemize}
    \item We introduce \our{}, a novel approach to the Few-Shot learning problem by aggregating information from the support set and directly producing weights updates.
    \item In \our{}, we do not use loss calculation or gradient backpropagation for the update to the new task, thus making the model more biologically feasible and computationally efficient.
    \item We significantly increase the update ability compared to the classical MAML algorithm, as evidenced by the increased accuracy in numerous benchmarks we perform.
\end{itemize}


\section{Related Work}

The problem of Meta-Learning and Few-Shot learning \citep{hospedales2020metalearning,schmidhuber1992fast,bengio1992optimization} has received a growing amount of attention from the scientific community over the recent years, with the abundance of methods emerging as a result. They can be roughly categorized into several groups:

\paragraph{Model-based methods} aim to adapt to novel tasks quickly by utilizing mechanisms such as memory \citep{ravi2016optimization,santoro2016meta,mishra2018simple,zhen2020learning}, Gaussian Processes \citep{rasmussen2003gaussian, patacchiola2020bayesian, wang2021learning}, or generating fast weights based on the support set with set-to-set architectures \citep{qiao2017fewshot, bauer2017discriminative, han2018learning}. The fast weights approaches can be interpreted as using Hypernetworks \citep{ha2016hypernetworks} -- models which learn to generate the parameters of neural networks performing the designated tasks. 

Similarly, \our{} utilizes a Hypernetwork to generate weights updates for performing specific tasks. The key difference is that in \our{}, the Hypernetwork is not the sole source of model weights, but is used to generate the updates to the weights of the base model.

\paragraph{Metric-based methods} learn a transformation to a feature space where the distance between examples from the same class is small. The earliest examples of such methods are Matching Networks \citep{vinyals2016matching} and Prototypical Networks \citep{snell2017prototypical}. Subsequent works show that metric-based approaches can be improved by techniques such as learnable metric functions \citep{sung2018learning}, conditioning the model on tasks \citep{oreshkin2018tadam} or predicting the parameters of the kernel function to be calculated between support and query data with Hypernetworks \citep{sendera2022hypershot}. \our{} does not consider relations between the support and query examples during inference and therefore does not fall into the category of metric-based models.

\paragraph{Optimization-based methods}

such as MetaOptNet \citep{lee2019meta} are based on the idea of an optimization process over the support set within the Meta-Learning framework. Arguably, the most popular of this family of methods is Model-Agnostic Meta-Learning (MAML) \citep{finn2017model}, which inspired a multitude of research and numerous extensions to the original algorithm. This includes various techniques for stabilizing its training and improving performance, such as Multi-Step Loss Optimization, and scheduling the learning rate of the meta-optimizer \citep{antioniou2018howto}
, or using the Bayesian variant of MAML \citep{yoon2018bayesian}.

Due to a need for calculating second-order derivatives when computing the gradient of the meta-training loss, training the classical MAML introduces a significant computational overhead. The authors show that in practice the second-order derivatives can be omitted at the cost of small gradient estimation error and minimally reduced accuracy of the model \citep{finn2017model, nichol2018first}. Methods such as iMAML and Sign-MAML propose to solve this issue with implicit gradients or Sign-SGD optimization \citep{rajeswaran2019meta, fan2021signmaml}.
The optimization process can also be improved by training not only the base initialization of the model but also the optimizer itself -- namely, training a neural network that transforms gradients calculated w.r.t. loss of the support set predictions into weight updates \citep{munkhdalai2017meta, munkhdalai2018rapid, li2017metasgd, rajasegaran2020pamela}.

\our{} shares a key characteristic with the optimization-based methods -- namely, it also utilizes a base set of weights, which are updated to obtain a model fit for a given task. The key difference between \our{} and MAML is that while MAML adapts to novel tasks through multiple steps of gradient-based optimization, \our{} generates the updates in a single step using a Hypernetwork. This makes \our{} more similar to methods like \citep{li2017metasgd, munkhdalai2017meta}, which generate weight updates through trained meta-optimizers. However, contrary to those approaches, in \our{} the Hypernetwork predicts the weight updates based on (i) latent representation of the support set, (ii) predictions of the base model for the support set, (iii) ground-truth labels of the support examples (see Fig. \ref{fig:architecture}). Thus \our{} does not require calculating either the loss function or its gradients during generating of the task-specific weight updates, making it more computationally efficient.

\section{\our{}: Hypernetwork for Few-Shot learning}

In this section, we present our \our{} model for Few-Shot learning. First, we start by presenting background and notations for Few-Shot learning. Then we describe how the MAML algorithm works. Finally, we present \our{}, which can be understood as an extension of the classical MAML. 

\subsection{Background}

\paragraph{The terminology}
 describing the Few-Shot learning setup is dispersive due to the colliding definitions used in the literature. 
Here, we use the nomenclature derived from the Meta-Learning literature, which is the most prevalent at the time of writing.
Let $\mathcal{S} = \{ (\mathbf{x}_l, \mathbf{y}_l) \}_{l=1}^L$ be a support-set containing input-output pairs, with $L$ examples with the equal class distribution. In the \textit{one-shot} scenario, each class is represented by a single example, and $L=K$, where $K$ is the number of the considered classes in the given task. Whereas, for \textit{Few-Shot} scenarios, each class usually has from $2$ to $5$ representatives in the support set $\mathcal{S}$. 

Let $\mathcal{Q} = \{ (\mathbf{x}_m, \mathbf{y}_m) \}_{m=1}^M$ be a query-set (sometimes referred to in the literature as a target-set), with $M$ examples, where $M$ is typically one order of magnitude greater than $K$. For clarity of notation, the support and query sets are grouped in a task $\mathcal{T} = \{\mathcal{S}, \mathcal{Q} \}$. During the training stage, the models for Few-Shot applications are fed by randomly selected examples from training set $\mathcal{D} = \{\mathcal{T}_n\}^N_{n=1}$,  defined as a collection of such tasks. 

 During the inference stage, we consider task $\mathcal{T}_{*} = \{\mathcal{S}_{*}, \mathcal{X}_{*}\}$, where $\mathcal{S}_{*}$ is a support set with the known class values for a given task, and $\mathcal{X}_{*}$ is a set of query (unlabeled) inputs. The goal is to predict the class labels for query inputs $\mathbf{x} \in \mathcal{X}_*$, assuming support set $\mathcal{S}_{*}$ and using the model trained on $\mathcal{D}$.





\begin{algorithm}[t]
    \caption{ MAML - Model-Agnostic Meta-Learning   \citep{finn2017model} } 
    \label{alg_overview_1}
    \textbf{Require:} $\mathcal{D}=\{\mathcal{T}_n\}_{n=1}^N$: set of training tasks \\
    \textbf{Require:}  $\alpha$, $\beta$: step size hyper parameters
    \begin{algorithmic}[1]
            \State randomly initialize $\theta$
            \While{ not done }
                \State Sample batch of tasks $\mathcal{B}$ from $\mathcal{D}$
                \For{each task $\mathcal{T}_i=\{\mathcal{S}_i, \mathcal{Q}_i\}$  from batch $\mathcal{B}$}
                    \State Evaluate $\nabla_{\theta} \L_{S_i} (f_{\theta})$ with respect to examples from $\mathcal{S}_i$ given loss by \eqref{eq:loss}
                    \State Compute adapted parameters $ \theta'_i$ with gradient descent using formula given by eq. \eqref{eq:task_update} 
                \EndFor
                \State Update the global parameters of the model $\theta$ with formula given by eq. \eqref{eq:global_upate}. 
           \EndWhile    
    \end{algorithmic}
     \label{algorithm:MAML}
\end{algorithm}

\paragraph{Model-Agnostic Meta-Learning (MAML)} is one of the current standard algorithms for Few-Shot learning, which learns the parameters of a model so that it can adapt to a new task in a few gradient steps.
    
We consider a model represented by a function $f_{\theta}$ with parameters $\theta$. In the Few-Shot problems $f_{\theta}$ models  discriminative probabilities for the classes, $f_{\theta} (\mathbf{x})=p(\mathbf{y}|\mathbf{x},\theta)$. The standard MAML model is trained with the procedure given by Algorithm~\ref{algorithm:MAML}. In each of the training iterations the batch of tasks $\mathcal{B}$ is sampled from $\mathcal{D}$. Further, for each task $\mathcal{T}_i=\{\mathcal{S}_i, \mathcal{Q}_i\}$ from $\mathcal{B}$  MAML adapts the model’s parameters $\theta'_i$ that are specific for a given task. The actual values of  $\theta'_i$ are calculated using one or more gradient descent updates. In the simplest case of one gradient iteration, the parameters are updated as follows:
\begin{equation}
 \theta'_i = \theta - \alpha \nabla_{\theta} \L_{\mathcal{S}_i} (f_{\theta}),
  \label{eq:task_update}
 \end{equation}

where $\alpha$ is the step size that may be fixed as a hyperparameter or meta-learned, and the loss function for a set of observations $\mathcal{Z}$ is defined as $\L_{\mathcal{Z}}$ for the few shot scenario is represented as a simple cross-entropy:

\begin{equation}
 \L_{\mathcal{Z}} (f_{\theta}) = \sum_{(\mathbf{x}_{i,l}, \mathbf{y}_{i,l}) \in \mathcal{Z}} \sum_{k=1}^K - y_{i,l}^k \log{f_{\theta, k} (\mathbf{x}_{i,j})}, 
 \label{eq:loss}
 \end{equation}
 
where $f_{\theta, k} (\mathbf{x}_{i,j})$ denotes $k$-th output of the model $f_{\theta}$, for a given input $\mathbf{x}_{i,l}$, and $\mathbf{y}_{i,l}$ is corresponding class in one-hot coding. For simplicity of notation, we will consider one gradient update for the rest of this section, but using multiple gradient updates is a straightforward extension. After calculating the tasks-specific updates $ \theta'_i$ the general model parameters are trained by optimizing for the performance of $f_{\theta'_i}$ with respect to $\theta$ across tasks from batch $\mathcal{B}$. More concretely, the meta-objective used to train the general parameters of the models is as follows:
\begin{equation}
\L_{MAML} (f_{\theta}) = \sum_{\mathcal{T}_i \in \mathcal{B}} \L_{\mathcal{Q}_i}(f_{\theta'}) = \sum_{\mathcal{T}_i \in \mathcal{B}} \L_{\mathcal{Q}_i} (f_{\theta - \alpha \nabla_{\theta} \L_{\mathcal{S}_i} (f_{\theta})}),
\label{eq:global_loss}
\end{equation}
 
 Note that the meta-optimization is performed over the model parameters $\theta$, whereas the objective is computed using the updated model parameters $\theta'$. In effect, our proposed method aims to optimize the model parameters such that one or a small number of gradient steps on a new task
will produce maximally effective behavior on that task.

The meta-optimization across tasks is performed via stochastic gradient descent (SGD) such that the model parameters $\theta$ are updated as follows:
\begin{equation}
\theta \leftarrow  \theta - \beta \nabla_{\theta} \L_{MAML} (f_{\theta}), 
\label{eq:global_upate}
\end{equation}

where $\beta$ is the meta step size.

During the inference stage, in order to perform predictions for newly observed task $\mathcal{T}_{*} = \{\mathcal{S}_{*}, \mathcal{X}_{*}\}$ the loss function  $\L_{\mathcal{S}_*} (f_{\theta})$ is calculated first using eq. \eqref{eq:loss}. Next, the parameters  $\theta'_*$ for task $\mathcal{T}_*$ are calculated from eq. \eqref{eq:task_update}. The final predictions for query examples $\mathcal{X}_*$ are performed by the model $f_{\theta'_*}$, where for selected query example $\mathbf{x}_q \in \mathcal{X}_*$ we have $p(\mathbf{y}|\mathbf{x}_q, \theta'_*)=f_{\theta'_*}(\mathbf{x}_q, \theta'_*)$. 

The main limitation of the approach is that it produces the general weights for all possible tasks, and the adjustment is performed via a gradient-based approach performed on the support set. For some non-trivial challenging tasks, the dedicated parameters $\theta'_*$ may be located far from the base weights, $\theta$. Consequently, the adaptation procedure would require significantly more gradient steps, the training may be unstable, and the model will tend to overfit to support set. To overcome this limitation, we propose to replace the gradient-based adaption with the Hypernetwork approach, where the update step is returned by an additional deep model that extracts the information from the support set (see toy example in Fig. \ref{fig:2D}).  






\subsection{\our{} - overview}

We introduce our {\our{}} -- a model that utilizes Hypernetworks for modeling weights updates in the classical MAML algorithm. The main idea of the proposed updating scenario is to use information extracted from support examples and predictions given by universal weights to find optimal updates dedicated to a given task. Thanks to this approach, we can switch the classifier's parameters between completely different tasks based on the support set and existing prediction of the universal weights. 

\begin{figure}[h!]
    \centering
    \includegraphics[width=\textwidth]{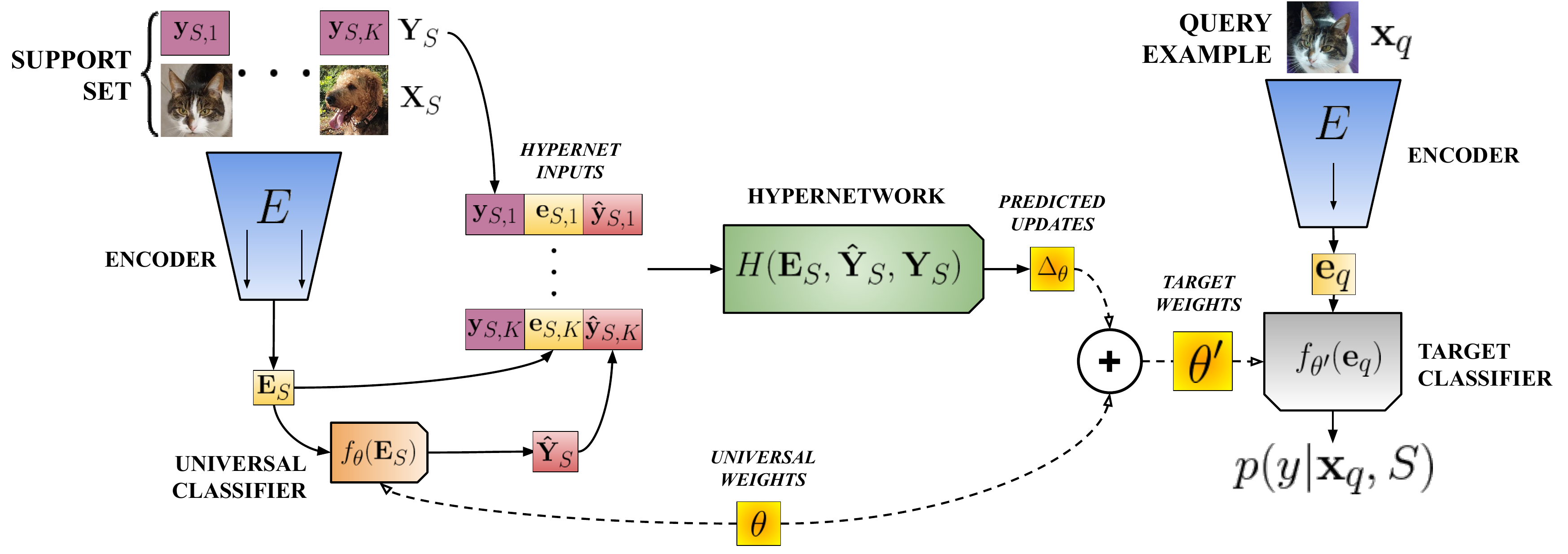}
    \caption{The overview of \our{} architecture. The input support examples are processed by encoding network $E(\cdot)$ and delivered to the hypernetwork $H(\cdot)$ together with the true support labels and predictions from general model $f_{\theta}(\cdot)$. The hypernetwork transforms them, and returns the update of weigths $\Delta \theta$ for target classifier $f_{\theta'}$. The query example is transformed by Encoder $E(\cdot)$, and the final class distribution is returned by the target model $f_{\theta'}$ dedicated to the considered task.  }
    \label{fig:architecture}
\end{figure}

The architecture of the \our{} is provided in Fig.~\ref{fig:architecture}. In this section we present the model for one-shot scenario, we further discuss, how to extend it to Few-Shot problems. We aim at predicting the class distribution $p(\mathbf{y}|\textbf{x}_q, \mathcal{S})$, assuming given single query example $\mathbf{x}_q$, and the set of support examples $\mathcal{S}$. Following the idea from MAML we consider the parameterized function $f_{\theta}$, that models the discriminative distribution for the classes. In addition, in our architecture we distinguish the trainable encoding network $E(\cdot)$, that transforms data to low-dimensional representation. We postulate to calculate $p(\mathbf{y}|\textbf{x}_q,\theta')=f_{\theta'}(\mathbf{e}_q)$, where $\mathbf{e}_q$ is the query example $\mathbf{x}_q$ transformed using encoder $E(\cdot)$, and $\theta'$ represents the updated parameters for a considered task, $\theta' = \theta + \Delta_{\theta}$. Compared to gradient-based adaptation step described by eq. \eqref{eq:task_update} used to calculate $ \Delta_{\theta}$ we propose to predict the values using hypernetwork, directly from support set.  

Each of the inputs from support set $\mathcal{X}_S$ is transformed by Encoder $E(\cdot)$ in order to obtain low-dimensional matrix of embeddings $\mathbf{E}_{\mathcal{S}}=[\mathbf{e}_{\mathcal{S},1}, \dots, \mathbf{e}_{\mathcal{S},K}]^{\mathrm{T}}$. Next, the corresponding class labels for support examples, $\mathbf{Y}_{\mathcal{S}}= [\mathbf{y}_{\mathcal{S},1}, \dots, \mathbf{y}_{\mathcal{S}, K}]^\mathrm{T}$ are concatenated to the corresponding embeddings stored in the rows of matrix $\mathbf{E}_{\mathcal{S}}$. In addition, we also calculate the predicted values for the examples from the support set using the general model $f_{\theta}(\mathbf{E}_{\mathcal{S}})=\mathbf{\hat{Y}_{\mathcal{S}}}$, and also concatenate them to $\mathbf{E}_{\mathcal{S}}$. The matrix transformed support inputs  $\mathbf{E}_{\mathcal{S}}$, together with true support labels $\mathbf{Y}_{\mathcal{S}}$, and corresponding predictions $\mathbf{\hat{Y}_{\mathcal{S}}}$ returned by genreal model are delivered as an input to the hypernetwork $H(\cdot)$ that returns the update $ \Delta_{\theta}$. The parameters for final target model are calculated with the following formula:

\begin{equation}
 \theta' = \theta  + \Delta_{\theta} = \theta + H( \mathbf{E}_{\mathcal{S}},\mathbf{\hat{Y}}_{\mathcal{S}}, \mathbf{Y}_{\mathcal{S}}).    
 \label{eq:hyper_update}
\end{equation}

Practically, the Hypernetwork observes the support examples with the corresponding true values and decides how the global parameters $\theta$ should be adjusted to the considered task. In addition, the predictions from global model $f_{\theta}$ are also delivered to the model in order to identify the misclassifications and try to correct them during the update state. 

\subsection{\our{} - training}

For training the model we assume that encoder $E(\cdot)$ is parametrized by $\gamma$, $E:=E_{\gamma}$, and the hypernetwork $H(\cdot)$ by $\eta$, $H:=H_{\eta}$. The training procedure is described in Algorithm~\ref{algorithm:HyperMAML}. First, we sample the batch of tasks $\mathcal{B}$ from the given dataset $\mathcal{D}$. Next, for each task $\mathcal{T}_i$ in batch $\mathcal{B}$ we calculate the update $\Delta_{\theta_i}$ using the support set $\mathcal{S}_i$, and provide the updated parameters $\theta'$ according to the rule given by eq. \eqref{eq:hyper_update}. Finally, the objective to train the parameters of the system is calculated using the query sets from the batch tasks $\mathcal{T}_i$:

\begin{equation}
\L_{HyperMAML} (f_{\theta}) = \sum_{\mathcal{T}_i \in \mathcal{B}} \L_{\mathcal{Q}_i}(f_{\theta'}) = \sum_{\mathcal{T}_i \in \mathcal{B}} \L_{\mathcal{Q}_i} (f_{\theta  + \Delta_{\theta}}),
\label{eq:hyper_global_loss}
\end{equation}
where $\L_{\mathcal{Q}_i}(f_{\theta'})$ is given by eq. \eqref{eq:loss}. The parameters of the encoder, hypernetwork, and global parameters $\theta$ represent the meta parameters of the system, and they are updated with stochastic gradient descent (SGD) by optimizing $\L_{HyperMAML} (f_{\theta})$. 

\begin{algorithm}[t]
    \caption{ \our{} } 
    \label{alg_overview_2}
    \textbf{Require:} $\mathcal{D}=\{\mathcal{T}_n\}_{n=1}^N$: set of training tasks \\
    \textbf{Require:}  $\beta$: step size hyper parameter
    \begin{algorithmic}[1]
            \State randomly initialize $\theta$, $\gamma$, $\eta$
            \While{ not done }
                \State Sample batch of tasks $\mathcal{B}$ from $\mathcal{D}$
                \For{each task $\mathcal{T}_i=\{\mathcal{S}_i, \mathcal{Q}_i\}$  from batch $\mathcal{B}$}
                    \State Compute adapted parameters $ \theta'_i$ from $\mathcal{S}_i$ using formula given by eq. \eqref{eq:hyper_update}
                \EndFor
                \State Calculate the loss $\L_{HyperMAML} (f_{\theta})$ given by eq. \eqref{eq:hyper_global_loss}.   
                \State $\theta \leftarrow  \theta - \beta \nabla_{\theta} \L_{HyperMAML} (f_{\theta})$ \Comment{Update the global target parameters $\theta$}
                \State $\eta \leftarrow  \eta - \beta \nabla_{\eta} \L_{HyperMAML} (f_{\theta})$ \Comment{Update parameters of the hypernetwork $H_{\eta}$}
                \State $\gamma \leftarrow  \gamma - \beta \nabla_{\gamma} \L_{HyperMAML}  (f_{\theta})$ \Comment{Update the parameters of the Encoder $E_{\gamma}$}
                \EndWhile
    \end{algorithmic}
     \label{algorithm:HyperMAML}
\end{algorithm}

 \paragraph{Adaptation to the Few-Shot scenario.} The proposed method can be easily extended for Few-Shot scenarios following the aggregation technique from \citep{sendera2022hypershot}. For our approach, we aggregate the embedding values of the support examples from the same class using \emph{mean} operation. In addition, the corresponding predictions within the class are also averaged and concatenated to the averaged per class embedding together with the true class label, and further processed via hypernetwork.

 \paragraph{Warming-up universal weights} 
In practice, it is not trivial to initialize the universal weights of \our{}. Classical initialization does not allow to update the universal weights and only the Hypernetwork's parameters are changing. To solve this problem we use a smooth transition from gradient to Hypernetwork update:  

\begin{equation}
 \theta' =  \theta + \lambda \cdot H( \mathbf{E}_{\mathcal{S}},\mathbf{\hat{Y}}_{\mathcal{S}}, \mathbf{Y}_{\mathcal{S}}) - (1 - \lambda) \cdot \alpha \nabla_{\theta} \L_{\mathcal{S}_i} (f_{\theta})    
\end{equation}

where $\lambda$ is changing from zero to one in a few initial training epochs.

\section{Experiments}

In this section, we first demonstrate the differences between MAML and \our{} on a toy 2D example showcased in Figure \ref{fig:2D}. Next, we compare \our{} to a variety of other Few-Shot learning approaches on classical Few-Shot benchmarks, as well as cross-domain Few-Shot learning. We next compare MAML and \our{} in terms of computational efficiency, as well as the magnitude of parameter updates. We conclude this section with an ablation study of two mechanisms we employ for training \our{}, namely switching and embedding enhancement, as well as its hyperparameters.

In the typical Few-Shot learning setting, making a valuable and fair comparison between proposed models is often complicated because of the existence of the significant differences in architectures and implementations of known methods. In order to limit the influence of the deeper backbone (feature extractor) architectures, we follow the unified procedure proposed by \citep{chen2019closer}\footnote{Our code is available at \url{https://github.com/gmum/few-shot-hypernets-public}.}.


In all of the experiments reported in sections \ref{sec:classification}-\ref{sec:efficiency}, the tasks consist of 5 classes (5-way) and 1 or 5 support examples (1 or 5-shot). Unless indicated otherwise, all compared models use a known and widely utilized backbone consisting of four convolutional layers (each consisting of a 2D convolution, a batch-norm layer, and a ReLU non-linearity; each layer consists of 64 channels) and have been trained from scratch \citep{chen2019closer}. In all experiments, the query set of each task consists of 16 samples for each class (80 in total). We split the datasets into the standard train, validation, and test class subsets, used commonly in the literature \citep{ravi2016optimization, chen2019closer, patacchiola2020bayesian}. For each setting, we report the performance of \our{} averaged over three training runs. We provide the additional training details in the Appendix.




\subsection{2D example}\label{sec:2d}

To visualize the fact that the MAML algorithm is significantly limited by the weight update procedure based on the gradient method, we will analyze a simple 2D example. Let us consider a data set consisting of 4 Gaussian data; see the first column in Fig.~\ref{fig:2D}. In the meta-learning scenario, we produce a task that consists of samples from two horizontal or vertical ellipses with permuted labels. In practice, we would like to force a method to adapt to four tasks presented in the second to the fifth column in Fig.~\ref{fig:2D}. In the training procedure, we use a neural network with one layer -- in practice, a linear model. We can visualize weights and decision boundaries thanks to using such simple architecture. We marked the universal weights with the blue color and the updated parameters -- with red. 

The MAML model with one gradient step in the inner loop (see the first row in Fig.~\ref{fig:2D}) cannot update all four tasks. On the other hand, when we use five gradient updates, the model gives a reasonable solution (see the second row in Fig.~\ref{fig:2D}). In practice, for MAML, we cannot effectively use many steps in inner lop in the case of really high dimensional data. 

Our \our{} can solve the task by using the hypernetwork paradigm. We use only one update in our method, which can dramatically change weights. Notice that even for five steps, MAML cannot significantly update weight. In practice, it uses only rotations. 

\subsection{Classification}\label{sec:classification}
We next consider the classical Few-Shot learning scenario. 
We benchmark the performance of the \our{} and other methods on two challenging and widely considered datasets: Caltech-USCD Birds ({CUB}) \citep{wah2011cub} and {mini-ImageNet} \citep{ravi2016optimization}. 
{In the case of the mini-ImageNet dataset, the models are initialized with pretrained weights, following \citep{qiao2017fewshot}}. We compare \our{} to several MAML-related and Hypernetwork-based methods, as well as the current state-of-the-art algorithms in the tasks of 1-shot and 5-shot classification, and report the results in Table \ref{tab:conv45shotcubminiimagenet}\footnote{{We report the performance of baseline methods as measured with our codebase~\cite{chen2019closer,patacchiola2020bayesian} wherever possible, and otherwise as reported in the works they were introduced in.} 
We include an expanded version of Table \ref{tab:conv45shotcubminiimagenet} with a comparison to a wider selection of models, as well a comparison of models using a larger ResNet-12 backbone in the Appendix.}.

In the $1$-shot scenario, \our{} yields top-performing results ($66.11\%$) on the {CUB} dataset. On the {mini-ImageNet} dataset, \our{} is also the second-best performing approach, achieving an accuracy of $55.91 \%$, second only to DFSVLwF~\citep{gidaris2018dynamic} ($56.20 \%)$. In the $5$-shot setting, \our{} is among the top three best-performing models achieving both on the {CUB} and {mini-ImageNet} datasets, achieving $78.89\%$ and $71.72 \%$ accuracies, respectively, with FEAT, {HyperShot \citep{sendera2022hypershot}, and Unicorn-MAML~\cite{ye2021unicorn} outperforming it by small margins.}
The obtained results show that \our{} achieves performance better or comparable to a variety of Few-Shot learning methods, in particular MAML \citep{finn2017model}, as well as techniques which derive from it \citep{antioniou2018howto,rajeswaran2019meta,fan2021signmaml,yoon2018bayesian,ye2021unicorn}.

\begin{table}[h!]

\centering
\caption{The classification accuracy results for the inference tasks on $\textbf{CUB}$ and $\textbf{mini-ImageNet}$ datasets in the $1$-shot and $5$-shot settings. The highest results are in bold and the second-highest in italics (the larger, the better). }
\label{tab:conv45shotcubminiimagenet}
{\scriptsize
\begin{tabular}{@{}l@{}c@{\;}c@{\;}c@{\;}c@{}}
\toprule
& \multicolumn{2}{c}{\textbf{CUB}}  & \multicolumn{2}{c}{\textbf{mini-ImageNet}} \\ 
\textbf{Method} & 1-shot & 5-shot  & 1-shot & 5-shot \\
\midrule

\textbf{MAML} \citep{finn2017model} & {$55.92 \pm 0.95$} & {$72.09 \pm 0.76$} &  {$48.70 \pm 1.75$}  & {$63.11 \pm 0.92$} \\
\textbf{MAML++} \citep{antioniou2018howto}   & -- & -- & $52.15 \pm 0.26 $ & $68.32 \pm 0.44 $\\
\textbf{iMAML-HF} \citep{rajeswaran2019meta} & --  & -- & $49.30 \pm 1.88$  & --\\
\textbf{SignMAML} \citep{fan2021signmaml} & -- & -- & $42.90 \pm 1.50$ & $60.70 \pm 0.70$ \\
\textbf{Bayesian MAML} \citep{yoon2018bayesian}  &  -- & -- & ${53.80 \pm 1.46}$  & $64.23 \pm 0.69$  \\
\textbf{Unicorn-MAML} \citep{ye2021unicorn} & $65.85 \pm 0.48$ & $76.63 \pm 0.39$ &  ${55.70} \pm 0.20 $ &  $\mathbf{72.68 \pm 0.16}$ \\

\textbf{HyperShot} \citep{sendera2022hypershot} & $65.27 \pm 0.24$  & $\mathit{79.80 \pm 0.16}$ & $52.42 \pm 0.46$  & ${68.78 \pm 0.29}$ \\
\textbf{PPA} \citep{qiao2017fewshot} & --  & -- & $54.53 \pm 0.40$ & -- \\

\textbf{DFSVLwF} \citep{gidaris2018dynamic} & -- & -- &  $\mathbf{56.20 \pm 0.86}$  & -- \\

\midrule

\textbf{\our{}} & $\mathit{66.11 \pm 0.28} $ & $ 78.89 \pm 0.19 $ & $\mathit{55.91 \pm 0.21}$ & $\mathit{71.72 \pm 0.16}$ \\ 
\bottomrule
\end{tabular}
}

\end{table}

\begin{table}
\caption{The classification accuracy results for the inference tasks on cross-domain tasks (\textbf{Omniglot}$\rightarrow$\textbf{EMNIST} and \textbf{mini-ImageNet}$\rightarrow$\textbf{CUB}) datasets in the $1$-shot and $5$-shot setting. The highest results are in bold and the second-highest in italic (the larger, the better). }
\centering
{\scriptsize
\begin{tabular}{@{}l@{}cccc@{}}
\toprule
\textbf{} & \multicolumn{2}{c}{\textbf{Omniglot}$\rightarrow$\textbf{EMNIST}} & \multicolumn{2}{c}{\textbf{mini-ImageNet}$\rightarrow$\textbf{CUB}} \\
 {\textbf{Method}} & \textbf{1-shot}& \textbf{5-shot} & \textbf{1-shot} & \textbf{5-shot} \\ 
\midrule
 {\textbf{Feature Transfer}}  \citep{zhuang2020comprehensive} & 64.22 $\pm$  {1.24} & 86.10 $\pm$  {0.84} & 32.77 $\pm$  {0.35} & 50.34 $\pm$  {0.27}\\
 {\textbf{Baseline$++$}} \citep{chen2019closer} & 56.84 $\pm$  {0.91} & 80.01 $\pm$  {0.92} & 39.19 $\pm$  {0.12} & ${ 57.31 \pm 0.11 }$ \\
 {\textbf{MatchingNet}} \citep{vinyals2016matching}  & 75.01 $\pm$  {2.09} & 87.41 $\pm$  {1.79} & 36.98 $\pm$  {0.06} & 50.72 $\pm$  {0.36} \\
 {\textbf{ProtoNet}} \citep{snell2017prototypical} & 72.04 $\pm$  {0.82} & 87.22 $\pm$  {1.01} & 33.27 $\pm$  {1.09} & 52.16 $\pm$  {0.17} \\

 {\textbf{RelationNet}} \citep{sung2018learning} & 75.62 $\pm$  {1.00} & 87.84 $\pm$  {0.27}  & 37.13 $\pm$  {0.20} & 51.76 $\pm$  {1.48}\\
 {\textbf{DKT}} \citep{patacchiola2020bayesian} & 75.40 $\pm$  {1.10} & $\mathit{90.30 \pm  0.49}$ & $\mathbf{40.14 \pm  {0.18}}$ & 56.40 $\pm$  {1.34} \\

 \textbf{HyperShot } \citep{sendera2022hypershot} &  ${78.06 \pm 0.24}$ & $89.04 \pm 0.18 $ & $39.09 \pm 0.28$ & $\mathit{57.77 \pm 0.33}$ \\
\textbf{HyperShot + finetuning} & $\mathbf{80.65 \pm 0.30}$ & $\mathbf{90.81 \pm 0.16}$ & $\mathit{40.03 \pm 0.41 }$ & $\mathbf{58.86 \pm 0.38}$ \\
{\textbf{MAML}} \citep{finn2017model} & 74.81 $\pm$  {0.25} & 83.54 $\pm$  {1.79}  & 34.01 $\pm$  {1.25} &48.83 $\pm$  {0.62} \\
{\textbf{Unicorn-MAML} \citep{ye2021unicorn}} & $76.70 \pm 0.54$ & $86.44 \pm 0.39$ & $35.40 \pm 0.17$ & $48.90 \pm 0.16$ \\
\textbf{Bayesian MAML} \citep{yoon2018bayesian} & $63.94 \pm 0.47$ & $65.26 \pm 0.30 $ & $33.52 \pm 0.36 $ & $51.35 \pm 0.16$ \\

\midrule 

 \textbf{\our{}} & $\mathit{79.84 \pm 0.59} $ & $ {89.22 \pm 0.78}$ &  $37.62 \pm 0.17$ & $50.90 \pm 0.17$ \\ 
\bottomrule
\end{tabular}
}
\label{tab:crossdomain_accuracy}

\end{table}

\subsection{Cross-domain adaptation}\label{sec:cross_domain}

In the cross-domain adaptation setting, the model is evaluated on tasks coming from a different distribution than the one it had been trained on. 
To benchmark the performance of \our{} in cross-domain adaptation, we combine two datasets so that the training fold is drawn from the first dataset and validation and the testing fold -- from another one. 
We report the results in Table \ref{tab:crossdomain_accuracy}. 

In the task of 1-shot {Omniglot}$\rightarrow${EMNIST} classification, \our{} achieves the second-best result ($79.84\%$), with HyperShot+finetuning ($80.65\%$) being the top one. Compared to the other methods, we observe relatively smaller performance growth as more data becomes available. In the $5$-shot  {Omniglot}$\rightarrow${EMNIST} classification task \our{} yields comparable results ($89.22\%$) to HyperShot \citep{sendera2022hypershot} ($90.81\%$) and DKT \citep{patacchiola2020bayesian} ($90.30\%$), which are the state-of-the-art in this setting. In the most challenging task of {mini-ImageNet}$\rightarrow${CUB} classification, our method performs comparably to baseline methods such as MAML, Unicorn-MAML, ProtoNet, and Matching Net \citep{finn2017model,ye2021unicorn,vinyals2016matching,snell2017prototypical}, particularly in the $1$-shot setting.

\subsection{Computational efficiency}
\label{sec:efficiency}

\begin{wraptable}{r}{0.52\textwidth}
    \scriptsize
    \centering
    \vspace{-1cm}
        \caption{Time (in seconds) spent on processing the entire \textbf{Omniglot} $\rightarrow$ \textbf{EMNIST} test dataset (600 tasks) by MAML with different numbers of gradient steps and \our{}.}
    \label{tab:efficiency}
    \begin{tabular}{lccc}
\toprule
 \textbf{Model} &  \textbf{Steps} & \textbf{Time} & \textbf{Accuracy} \\
\midrule
      \multirow{9}{*}{MAML} &      0 &   $2.17 \pm 0.28$ & $19.98 \pm 0.35$ \\
      &      1 &   $5.16 \pm 0.13$ & $74.67 \pm 0.25$ \\
      &      2 &   $7.30 \pm 0.11$ & $74.69 \pm 0.24$ \\
       &      3 &   $8.87 \pm 0.08$ & $74.71 \pm 0.25$ \\
       &      4 &  $10.79 \pm 0.19$ & $74.71 \pm 0.24$ \\
       &      5 &  $12.66 \pm 0.14$ & $74.71 \pm 0.23$ \\
      &     10 &  $22.60 \pm 0.28$ & $74.74 \pm 0.25$ \\
       &     25 &  $51.75 \pm 0.24$ & $74.78 \pm 0.25$ \\
       &     50 & $103.88 \pm 2.24$ & $74.82 \pm 0.25$ \\
      &    100 & $212.62 \pm 2.24$ & $74.86 \pm 0.27$ \\
      \midrule
      \our{} &  -- &   $8.21 \pm 0.41$ & $79.84 \pm 0.59$ \\

\bottomrule
\end{tabular}

    \vspace{-0.5cm}
\end{wraptable}

Finally, we verify the hypothesis that \our{} offers an increased computational efficiency compared to MAML. To this end, we measure the times of processing the entire \textbf{Omniglot} $\rightarrow$ \textbf{EMNIST} test dataset (600 tasks in total) by MAML with different numbers of gradient steps and \our{} and report the results in Table \ref{tab:efficiency}. We provide additional results for other datasets in the supplementary material.
We find that processing the test data with \our{} takes approximately the same time as using MAML with just 2 gradient updates. We also note that even given the budget of 100 gradient updates, MAML never matches the accuracy achieved by a single update generated by \our{}.

\section{Conclusions}
\label{sec:conclusion}
In this work, we introduced \our{} -- a novel Meta-Learning algorithm strongly motivated by MAML \citep{finn2017model}. The crucial difference between the two methods lies in the fact that in \our{} the update of the model is given not by the gradient optimization, but by the trainable Hypernetwork. Consequently, \our{} is more computationally efficient than MAML, as during adaptation it performs only a single parameter update. 
Moreover, \our{} is not only more biologically feasible as it does not use backpropagation, but it can adapt easier to different settings and has a smaller number of hyperparameters. Our experiments show that  \our{} outperforms the classical MAML in a number of standard Few-Shot learning benchmarks and achieves results similar to various other state-of-the-art methods in most cases. 

\paragraph{Limitations} 
The main limitation of the method is the need for initialization of the universal weights by a continuous transition from gradient update to Hypernetwork. We plan to investigate this issue in future work.


\bibliographystyle{plain}

\appendix

\section{Ablation study of mechanisms used in \our{}}

In this section, we present two mechanisms we utilize when training \our{}.  

\paragraph{Switching} mechanism is a smooth transition between training the convolutional encoder through MAML and \our{} objective. We consider the MAML warm-up as a starting point of the training loop and then smoothly move towards \our{} training. During training, we define two "milestone" epochs (see Section \ref{app:sec:hyperparameters}, between which the transition occurs. During the transition, we continuously decrease the participation of the MAML objective in the training process. It is done by multiplying MAML loss by $p$, ranging from $1.0$ to $0.0$ for a given number of epochs, and multiplying the \our{} loss by $1-p$. Our motivation for this mechanism is to train a better universal optimizing the MAML objective during the warm-up part of the training and then Switch to the \our{} objective gradually.

\begin{figure}[h!]
    \centering
    \includegraphics[width=0.8\textwidth]{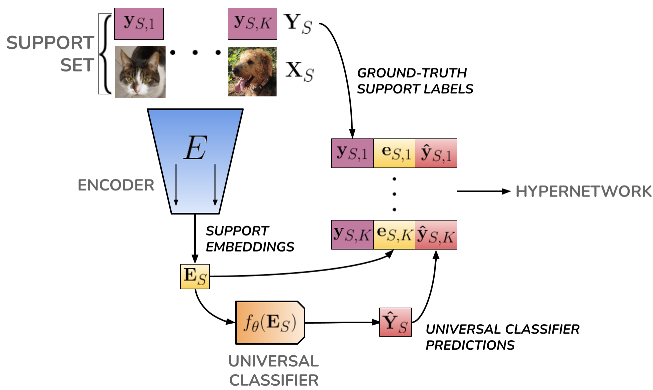}
    \caption{Illustration of the embedding enhancement mechanism. The support embeddings (which serve as the input to the \our{} Hypernetwork) are enhanced with the predictions of the base classifier and their respective ground-truth labels.}
    \label{fig:enhancement}
\end{figure}

\paragraph{Enhancement} of the embeddings is another mechanism in our framework. When preparing the input to the Hypernetwork from the support set, we first obtain support embeddings from the encoder. Then, we forward those embeddings through the universal classifier and obtain its predictions for support examples. We concatenate those predictions to the support embeddings, as well as their respective ground-truth labels. The whole process is visualized in Figure \ref{fig:enhancement}. Our motivation to perform such an operation is to give the Hypernetwork the information about the current decision of the classifier, to better estimate the updates to its weights. 
We note that even though the Hypernetwork generates the weights of the classifier based on the enhanced support embeddings, the generated downstream classifier does not use enhancements when processing the query set. Instead, it only processes the raw embeddings obtained from the encoder (see Figure 2 from the main paper).

We perform an ablation study of both mechanisms on the task 5-shot 1-way \textbf{Omniglot}$\rightarrow$\textbf{EMNIST} classification. The results, reported in Table \ref{tab:ablation}, indicate that both mechanisms utilized individually improve the performance of \our{}, and the combination of the two yields the best results.


\begin{table}[h!]
    \label{tab:ablation}
    \scriptsize
    \centering
        \caption{The classification accuracy of \our{} on \textbf{Omniglot}$\rightarrow$\textbf{EMNIST} task in the 5-way 1-shot scenario. We consider different strategies by adding switch or embedding enhancement strategy to our model.}
    \label{tab:efficiency_1}
    \begin{tabular}{lccccc}
\toprule
 \textbf{Model} & \textbf{Switch} & \textbf{Enhancement} & \textbf{Accuracy} & \textbf{Learning rate} & \textbf{Milestones} \\
\midrule
      \multirow{4}{*}{\our{}} & no & no & $ 74.39 \pm 0.88 $ & $0.0001$ & - \\
       & no & yes & $ 75.28 \pm 1.29 $ & $0.0001$ & - \\
       & yes & no & $ 76.89 \pm 0.71 $ & $0.01$ & $51, 550$ \\
       & yes & yes & $ 79.10 \pm 1.43 $ & $0.01$ & $51, 550$ \\

\bottomrule
\end{tabular}

\end{table}


\section{Full classification results}

In Table \ref{tab:conv4full} we provide an expanded version of Table \ref{tab:conv45shotcubminiimagenet}, with numerous additional baseline methods. We also report the performance of \our{}, and several baseline methods trained with ResNet-12  backbone on the \textbf{mini-ImageNet} dataset in Table \ref{tab:miniresnet12}. 

\begin{table}[h]
\centering
\caption{The classification accuracy results for the inference tasks on $\textbf{CUB}$ and $\textbf{mini-ImageNet}$ datasets in the $5$-shot setting. The highest results are in bold and the second-highest in italics (the larger, the better). }
\label{tab:conv4full}

{\scriptsize
\begin{tabular}{@{}l@{}c@{\;}c@{\;}c@{\;}c@{}}
\toprule
& \multicolumn{2}{c}{\textbf{CUB}}  & \multicolumn{2}{c}{\textbf{mini-ImageNet}} \\ 
\textbf{Method} & 1-shot & 5-shot  & 1-shot & 5-shot \\
\midrule
\textbf{ML-LSTM} \citep{ravi2016optimization} & --  & -- & $43.44 \pm 0.77$ & $60.60 \pm 0.71$ \\
\textbf{SNAIL} \citep{mishra2018simple} & -- & -- & $45.10$  & $55.20$ \\
\textbf{LLAMA} \citep{grant2018recasting} & --  & -- &$49.40 \pm 1.83$  \\
\textbf{VERSA} \citep{gordon2018meta} & -- &  -- & $48.53 \pm 1.84$  & $67.37 \pm 0.86$ \\
\textbf{Amortized VI} \citep{gordon2018meta} & -- & -- & $44.13 \pm 1.78$  & $55.68 \pm 0.91$ \\
\textbf{Meta-Mixture} \citep{jerfel2019reconciling} & -- & -- & $49.60 \pm 1.50$  & $64.60 \pm 0.92$ \\
\textbf{SimpleShot} \citep{wang2019simpleshot} & -- & -- & $49.69 \pm 0.19$ & $66.92 \pm 0.17$ \\
\textbf{Feature Transfer} \citep{zhuang2020comprehensive}  & $46.19 \pm 0.64$ &  $68.40 \pm 0.79$ & $39.51 \pm 0.23$ & $60.51 \pm 0.55$ \\
\textbf{Baseline++} \citep{chen2019closer} & $61.75 \pm 0.95$  & $78.51 \pm 0.59$ & $47.15 \pm 0.49$ & $66.18 \pm 0.18$ \\
\textbf{MatchingNet} \citep{vinyals2016matching} & $60.19 \pm 1.02$ & $75.11 \pm 0.35$ & $48.25 \pm 0.65$  & $62.71 \pm 0.44$ \\
\textbf{ProtoNet} \citep{snell2017prototypical} &  $52.52 \pm 1.90$ & $75.93 \pm 0.46$ & $44.19 \pm 1.30$ & $64.07 \pm 0.65$ \\
\textbf{RelationNet} \citep{sung2018learning} & $62.52 \pm 0.34$  & $78.22 \pm 0.07$  & $48.76 \pm 0.17$ & $64.20 \pm 0.28$ \\
\textbf{DKT + CosSim} \citep{patacchiola2020bayesian} & $63.37 \pm 0.19$  & $77.73 \pm 0.26$ & $48.64 \pm 0.45$  & $62.85 \pm 0.37$ \\
\textbf{DKT + BNCosSim} \citep{patacchiola2020bayesian} & $62.96 \pm 0.62$  & $77.76 \pm 0.62$  & $49.73 \pm 0.07$  & $64.00 \pm 0.09$ \\
\textbf{GPLDLA} \citep{kim2021gaussian} &  $63.40 \pm 0.14$ & $78.86 \pm 0.35$ & $52.58 \pm 0.19$ & -- \\
\textbf{VAMPIRE} \citep{nguyen2020uncertainty}& -- & -- & $51.54 \pm 0.74$ &  $64.31 \pm 0.74$ \\
\textbf{PLATIPUS} \citep{finn2018probabilistic} & -- & -- & $50.13 \pm 1.86$  & -- \\
\textbf{ABML} \citep{ravi2018amortized} & $49.57 \pm 0.42$  & $68.94 \pm 0.16$ & $45.00 \pm 0.60$ & -- \\
\textbf{OVE PG GP (ML)} \citep{snell2020bayesian}   & $63.98 \pm 0.43$ & $77.44 \pm 0.18$  & $50.02 \pm 0.35$ & $64.58 \pm 0.31$\\
\textbf{OVE PG GP (PL)}  \citep{snell2020bayesian}  & $60.11 \pm 0.26$ & $79.07 \pm 0.05$  & $48.00 \pm 0.24$ & $67.14 \pm 0.23$ \\
\textbf{Reptile} \citep{nichol2018first} & -- & -- & $49.97 \pm 0.32$ & $65.99 \pm 0.58$  \\
\textbf{R2-D2} \citep{bertinetto2018meta} & --  & -- & $48.70 \pm 0.60$ & $65.50 \pm 0.60$ \\
\textbf{VSM} \citep{zhen2020learning} & --  & -- & ${54.73 \pm 1.60}$ &  $68.01 \pm 0.90$ \\
\textbf{PPA} \citep{qiao2017fewshot} & --  & -- & $54.53 \pm 0.40$ & -- \\
\textbf{DFSVLwF} \citep{gidaris2018dynamic} & -- & -- &  $\mathbf{56.20 \pm 0.86}$  & -- \\
\textbf{HyperShot} \citep{sendera2022hypershot} & ${65.27 \pm 0.24}$  & $ 79.80 \pm 0.16$ & $52.42 \pm 0.46$  & $68.78 \pm 0.29$ \\
\textbf{HyperShot+ finetuning} \citep{sendera2022hypershot} & $\mathit{66.13 \pm 0.26}$& $\mathit{80.07 \pm 0.22}$ & $53.18 \pm 0.45$  & $69.62 \pm 0.28$ \\
\textbf{MAML} \citep{finn2017model} & $56.11 \pm 0.69$ & $74.84 \pm 0.62$ &  $45.39 \pm 0.49$  & $61.58 \pm 0.53$ \\
\textbf{MAML++} \citep{antioniou2018howto}   & -- & -- & $52.15 \pm 0.26 $ & $68.32 \pm 0.44 $\\
\textbf{iMAML-HF} \citep{rajeswaran2019meta} & --  & -- & $49.30 \pm 1.88$  & --\\
\textbf{SignMAML} \citep{fan2021signmaml} & -- & -- & $42.90 \pm 1.50$ & $60.70 \pm 0.70$ \\
\textbf{Bayesian MAML} \citep{yoon2018bayesian}  & -- & -- & $53.80 \pm 1.46$  & $64.23 \pm 0.69$  \\
\textbf{Unicorn-MAML} \citep{ye2021unicorn} & $65.85 \pm 0.48$ & $76.63 \pm 0.39$ &  ${55.70} \pm 0.20 $ &  $\mathbf{72.68 \pm 0.16}$ \\

\textbf{Meta-SGD} \citep{li2017metasgd} & -- & -- & $50.47 \pm 1.87$ & $64.03 \pm 0.94$ \\
\textbf{MetaNet} \citep{munkhdalai2017meta} & -- & -- &  ${49.21 \pm 0.96}$  & -- \\
\textbf{PAMELA} \citep{rajasegaran2020pamela} & -- & -- & $53.50 \pm 0.89$ & ${70.51 \pm 0.67}$ \\

\midrule

\textbf{\our{}} & $\mathit{66.11 \pm 0.28} $ & $ 78.89 \pm 0.19 $ & $\mathit{55.91 \pm 0.21}$ & $\mathit{71.72 \pm 0.16}$ \\ 

\bottomrule
\end{tabular}
}
\end{table}

\begin{table}
\centering
\caption{The classification accuracy results for the inference tasks in the \textbf{mini-ImageNet} dataset in the 5-way (1-shot and 5-shot) scenarios. We consider models using the ResNet-12 backbone. The highest results are in bold and the second-highest in italics (the larger, the better). }
\label{tab:miniresnet12}
{\scriptsize
\begin{tabular}{@{}l@{}ccc@{}}
\toprule
\textbf{Method} &  \textbf{1-shot} & \textbf{5-shot} \\ \midrule
\textbf{cosine classifier}~\citep{chen2019closer}  & 55.43 $\pm$ 0.81 & 77.18 $\pm$ 0.61 \\
\textbf{TADAM}~\citep{oreshkin2018tadam}  & 58.50 $\pm$ 0.30 & 76.70 $\pm$ 0.30 \\
\textbf{HyperShot}~\citep{sendera2022hypershot} &  $59.12 \pm 0.26$ &  $76.53 \pm 0.22$  \\  
\textbf{HyperShot + adaptation}~\citep{sendera2022hypershot} & $60.30 \pm 0.31$ & $76.21 \pm 0.20$  \\ 
\textbf{ProtoNet}~\citep{snell2017prototypical}   & 62.39 $\pm$ 0.21  & 80.53 $\pm$ 0.14 \\ %
\textbf{MetaOptNet}~\citep{lee2019meta}  & 62.64 $\pm$ 0.82 & 78.63 $\pm$ 0.46\\
\textbf{MatchingNet}~\citep{vinyals2016matching}   & 63.08 $\pm$ 0.80 & 75.99 $\pm$ 0.60\\
\textbf{MAML}~\citep{finn2017model} & $63.11 \pm 0.92$ & 80.81 $\pm$ 0.14 \\
\textbf{PPA}~\citep{qiao2017fewshot} & $59.60$ & $73.34$ \\
%
\midrule

\textbf{\textbf{\our{}}} & 60.93 $\pm$ 0.22  & 75.74 $\pm$ 0.17 \\

\bottomrule 

\end{tabular}
}
\end{table}

\section{Computational efficiency -- results on natural image datasets}

In this section, we perform similar experiments to those described in Section~\ref{sec:efficiency} of the main paper and measure the inference time of \our{} and MAML with different numbers of gradient steps on \textbf{CUB} and \textbf{mini-ImageNet} datasets. We summarize the results in Table \ref{tab:efficiency_2}. Similarly to the benchmark performed on smaller images from the \textbf{Omniglot} $\rightarrow$ \textbf{EMNIST} dataset, the inference time of \our{} is comparable to MAML with two gradient steps. Likewise, MAML never achieves accuracy higher than \our{}.

As opposed to Section~\ref{sec:efficiency} of the main paper, we report the accuracies of MAML only up to seven gradient steps. This is due to an insufficient amount of GPU memory available for making more steps in the MAML implementation we used \citep{chen2019closer}. We also note that the accuracies of MAML reported here are significantly higher than the ones reported in Table 1 of the main paper. The MAML accuracies from that table were previously reported in the literature \citep{patacchiola2020bayesian}, whereas the results in Table \ref{tab:efficiency_2} have been obtained by the MAML implementation in our codebase \citep{chen2019closer}.

\begin{table}[h!]
    \centering
    \footnotesize
        \caption{Time (in seconds) spent on processing the entire \textbf{CUB} and \textbf{mini-ImageNet} test datasets (600 tasks each) by MAML with different numbers of gradient steps and \our{}.}
    \label{tab:efficiency_2}
    \begin{tabular}{lccccc}
\toprule
& & \multicolumn{2}{c}{\textbf{CUB}} & \multicolumn{2}{c}{\textbf{mini-ImageNet}} \\
 \textbf{Model} &  \textbf{Steps}  & \textbf{Time} & \textbf{Accuracy} & \textbf{Time} & \textbf{Accuracy} \\
\midrule
      \multirow{8}{*}{MAML} &  0 & $4.15 \pm 0.03$ & $20.01 \pm 0.02$ & $4.24 \pm 0.14$ & $20.10 \pm 0.31$ \\
      &      1 &  $6.89 \pm 0.10$ & $47.99 \pm 0.22$ &  $8.48 \pm 0.14$ & $43.46 \pm 0.42$ \\
       &      2 &  $9.44 \pm 0.15$ & $58.53 \pm 0.55$ & $10.50 \pm 0.80$ & $48.71 \pm 0.19$ \\
       &      3 & $11.60 \pm 0.02$ & $61.54 \pm 0.41$ & $13.17 \pm 0.26$ & $48.32 \pm 0.38$ \\
       &      4 & $12.97 \pm 0.09$ & $62.18 \pm 0.38$ & $15.17 \pm 0.18$ & $48.82 \pm 0.25$ \\
       &      5 &  $15.00 \pm 0.17$ & $62.66 \pm 0.40$ &  $17.29 \pm 0.32$ & $48.89 \pm 0.31$ \\
       &      6 & $16.90 \pm 0.12$ & $62.77 \pm 0.36$ & $19.01 \pm 0.26$ & $48.80 \pm 0.24$ \\
      &      7 & $19.07 \pm 0.17$ & $62.83 \pm 0.34$ & $20.55 \pm 0.30$ & $48.87 \pm 0.29$ \\
      \our{}  & -- & $8.46 \pm 0.37$ & $65.79 \pm 0.88$ & $11.11 \pm 0.25$ & $51.38 \pm 0.37$ \\

\bottomrule
\end{tabular}

\end{table}

\section{Training details}

In this section, we present details of the training and architecture overview.

\subsection{Architecture overview}

The architecture of \our{} consists of the following parts (as outlined in the Fig.~\ref{fig:architecture} in the main body of the work):

\paragraph{Encoder}

For each experiment described in the main body of this work, we utilize a shallow convolutional encoder (feature extractor), commonly used in the literature \citep{finn2017model, chen2019closer, patacchiola2020bayesian}.  This encoder consists of four convolutional layers, each consisting of a convolution, batch normalization, and ReLU nonlinearity. Each of the convolutional layers has an input and output size of 64, except for the first layer, where the input size is equal to the number of image channels. We also apply max-pooling between each convolution, by which the resolution of the processed feature maps is decreased by half. The output of the encoder is flattened to process it in the next layers. 

In the case of the \textbf{Omniglot} and \textbf{EMNIST} images, such encoder compresses the images into 64-element embedding vectors, which serve as input to the Hypernetwork (with the above-described enhancements) and the classifier. However, in the case of substantially larger \textbf{mini-ImageNet} and \textbf{CUB} images, the backbone outputs a feature map of shape $[64 \times 5 \times 5]$, which would translate to $1600$-element embeddings and lead to an over parametrization of the Hypernetwork and classifier which processes them and increase the computational load. Therefore, we apply an average pooling operation to the obtained feature maps and ultimately also obtain embeddings of shape $64$. Thus, we can use significantly smaller Hypernetworks.

\paragraph{Hypernetwork} 
The Hypernetwork transforms the enhanced embeddings of the support examples of each class in a task into the updates for the portion of classifier weights predicting that class. It consists of three fully-connected layers with ReLU activation function between each consecutive pair of layers. In the hypernetwork, we use a hidden size of $256$ or $512$.

\paragraph{Classifier} The universal classifier is a single fully-connected layer with the input size equal to the encoder embedding size (in our case 64) and the output size equal to the number of classes. When using the strategy with embeddings enhancement, we freeze the classifier to get only the information about the behavior of the classifier, this means we do not calculate the gradient for the classifier in this step of the forward pass. Instead, gradient calculation for the classifier takes place during the classification of the query data.

\subsection{Training details}

In all of the experiments described in the main body of this work, we utilize the switch and the embedding enhancement mechanisms. During training, we use the Adam optimizer and the \href{https://pytorch.org/docs/stable/generated/torch.optim.lr_scheduler.MultiStepLR.html}{MultiStepLR} learning rate scheduler with the decay of $0.3$ and learning rate starting from $0.01$ or $0.001$. We train \our{} for 4000 epochs on all the datasets, save for the simpler \textbf{Omniglot} $\rightarrow$ \textbf{EMNIST} classification task, where we train for 2048 epochs instead.

\subsection{Hyperparameters} 
\label{app:sec:hyperparameters}
{In this section, we present details of the model architecture, training procedure, as well as implementation. We summarize the hyperparameters of architecture and training procedures used in each experiment in Table \ref{tab:hyperparameters_1shot}.} 


\begin{table}[h]
    \caption{Hyperparameters for experiments conducted on each of the datasets.}
    \label{tab:hyperparameters_1shot}
    \scriptsize
    \centering
    \begin{tabular}{lcccc}
\toprule
\multirow{1}{*}{\textbf{hyperparameter}}
  & \, \multirow{1}{*}{\textbf{CUB}} \, & \, \multirow{1}{*}{\textbf{mini-ImageNet}} \, & \, \textbf{mini-ImageNet $\rightarrow$ CUB} \, & \, \textbf{Omniglot $\rightarrow$ EMNIST}  \\
\midrule
      learning rate & $0.001$ & $0.0001$ & $0.0001$ & $0.01$ \\
      Hypernetwork depth & $3$ & $3$ & $3$ & $3$ \\
      Hypernetwork width & $512$ & $512$ & $512$ & $512$ \\
      epochs no. & $4000$ & $4000$ & $4000$ & $2048$ \\
      switching milestones & $101, 1100$ & $101, 1100$  & $101, 1100$ & $51, 550$ \\
\bottomrule
\end{tabular}

\end{table}

\section{Implementation details}

We implement \our{} using the PyTorch framework. We attach the zipped folder containing the source code with our experiments as a part of the supplementary material. We shall release the code publicly after the end of the review period.

Each experiment described in this work was run on a single NVIDIA RTX 2080 GPU. Typical \our{} model trained on the \textbf{Omniglot} $\rightarrow$ \textbf{EMNIST} and \textbf{mini-ImageNet} datasets takes about 2 and 100 hours, respectively.

\end{document}